\documentclass[letterpaper, 10pt, conference]{ieeeconf}      

\IEEEoverridecommandlockouts                              

\usepackage{amsmath} 
\usepackage{amssymb}  
\usepackage{soul}

\def\MujocoAdaptive{
\centering{
\includegraphics[width=0.24\textwidth]{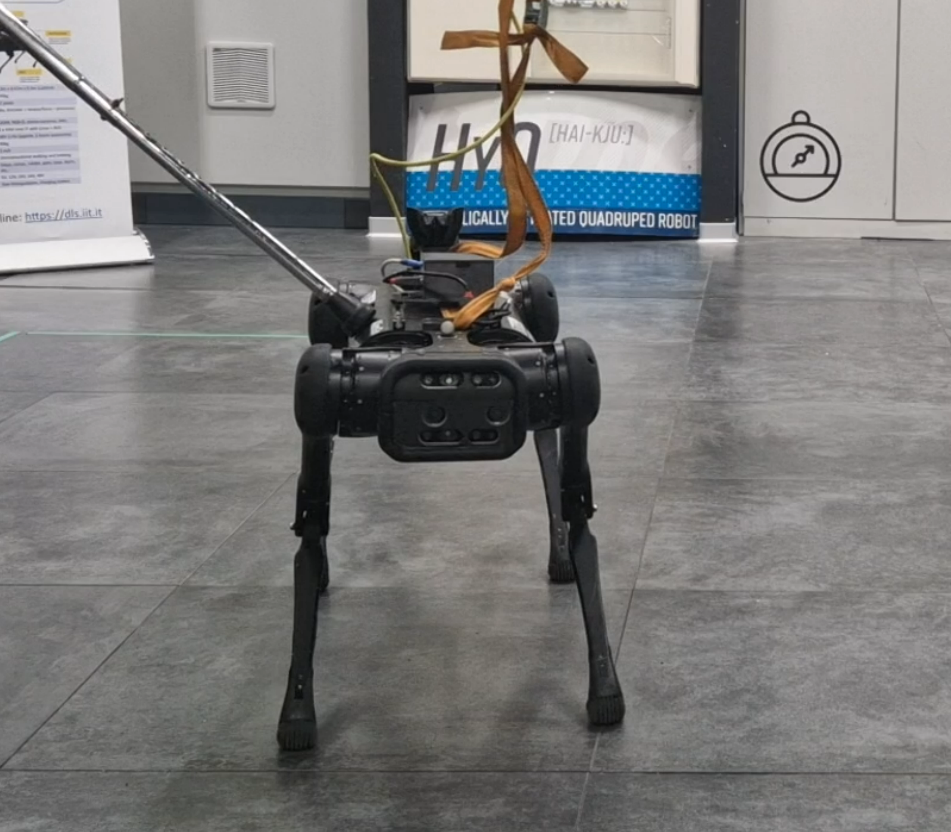}
\includegraphics[width=0.24\textwidth]{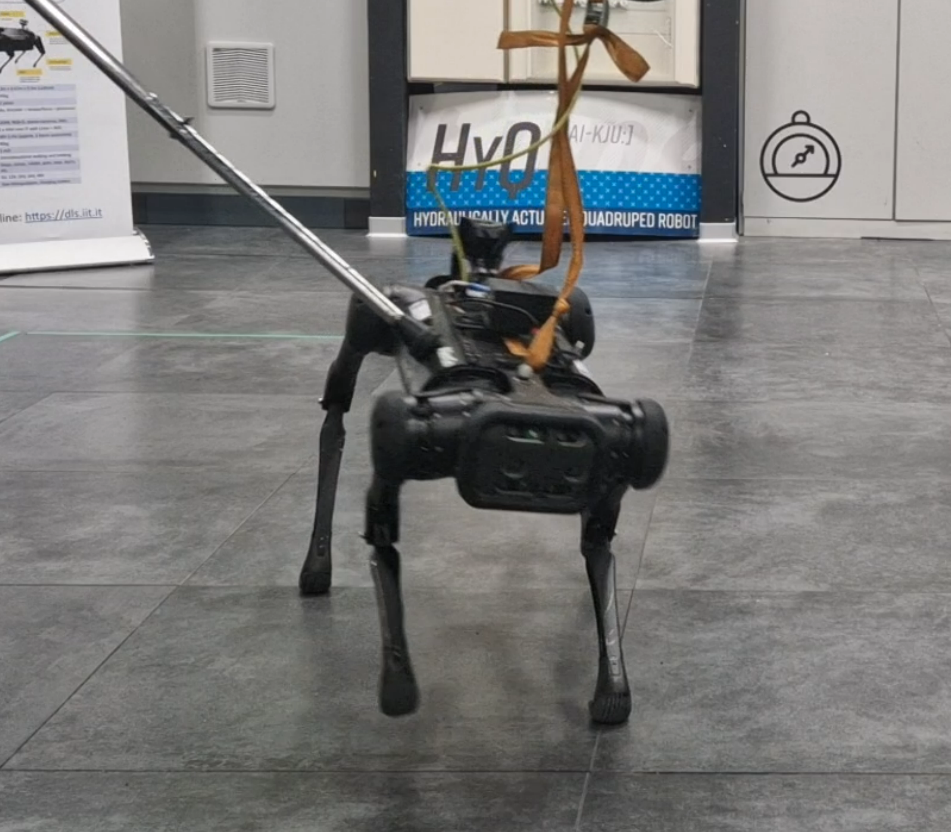}
\includegraphics[width=0.24\textwidth]{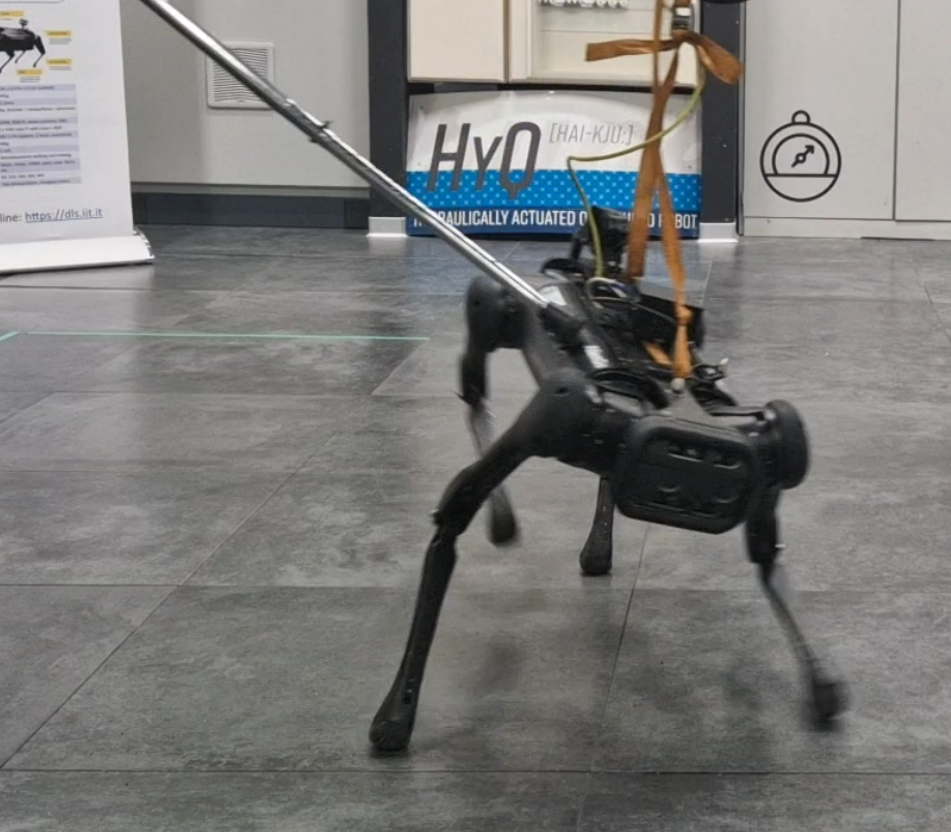}
\includegraphics[width=0.24\textwidth]{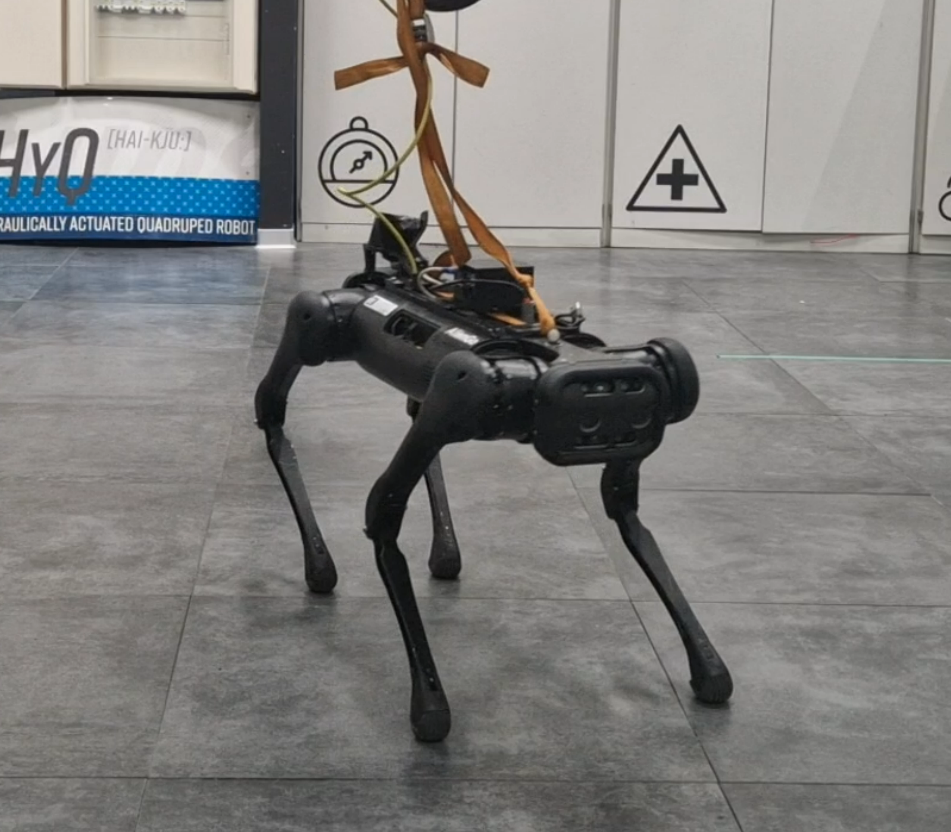}
}

}

\usepackage{xcolor}
\usepackage{eso-pic}
\AddToShipoutPictureBG*{%
  \AtPageUpperLeft{%
    \setlength{\unitlength}{1mm}%
    \put(-9.5,-9){\makebox(\paperwidth,0)[c]{\parbox{0.9\textwidth}{2024 IEEE/RSJ International Conference on Intelligent Robots and Systems (IROS), pp. 13757-13764}}}
  }
}

\AddToShipoutPictureBG*{%
  \AtPageUpperLeft{%
    \setlength{\unitlength}{1mm}%
    \put(-9.5,-14){\makebox(\paperwidth,0)[c]{\parbox{0.9\textwidth}{DOI: \href{https://ieeexplore.ieee.org/document/10801698}{10.1109/IROS58592.2024.10801698}}}}
  }
}

\title{\LARGE \bf
On the Benefits of GPU Sample-Based Stochastic \\ Predictive Controllers for Legged Locomotion
}

\author{Giulio Turrisi$^{1}$, Valerio Modugno$^{2}$, Lorenzo Amatucci$^{1}$, Dimitrios Kanoulas$^{2}$, Claudio Semini$^{1}$
\thanks{$^{1}$ These authors are with the Dynamic Legged Systems Laboratory, Istituto
Italiano di Tecnologia (IIT), Genova, Italy. E-mail: {\tt\small name.lastname@iit.it}}%
\thanks{$^{2}$ These authors are with the Department of Computer Science, University College London, Gower Street, WC1E 6BT, London, UK. E-mail: {\tt\small \{v.modugno, d.kanoulas\}@ucl.ac.uk}}%
\thanks{This work was partially supported by the UKRI Future Leaders Fellowship [MR/V025333/1] (RoboHike)}
}

\usepackage{amsmath}

\newcommand{\vect}[1]{\boldsymbol{#1}}
\newcommand{\mat}[1]{\boldsymbol{#1}}

\newcommand{\diffs}[3]{\frac{\partial^2 #1}{
\ifx#2#3 
\partial #2^2
\else
\partial #2 \partial #3
\fi
}}

\newcommand{\fv}{\vect{f}}
\newcommand{\gv}{\vect{g}}

\newcommand{\pv}{\vect{p}}

\newcommand{\qv}{{\vect{q}}}

\newcommand{\uv}{\vect{u}}
\newcommand{\vv}{\vect{v}}

\newcommand{\xv}{\vect{x}}

\newcommand{\deltav}{\vect{\delta}}

\newcommand{\sigmav}{\vect{\sigma}}

\newcommand{\piv}{\vect{\pi}}

\newcommand{\tauv}{\vect{\tau}}

\newcommand{\thetav}{\vect{\theta}}

\newcommand{\omegav}{\vect{\omega}}

\newcommand{\IIm}{\mat{I}}

\newcommand{\Cm}{\mat{C}}

\newcommand{\Em}{\mat{E}}

\newcommand{\Jm}{\mat{J}}

\newcommand{\Qm}{\mat{Q}}
\newcommand{\Rm}{\mat{R}}

\newcommand{\Gammam}{\mat{\Gamma}}
\newcommand{\Phim}{\mat{\Phi}}

\usepackage{duckuments}
\usepackage{graphicx}
\usepackage{amsmath}
\usepackage{algpseudocode}
\usepackage{algorithm}
\usepackage{xcolor}
\usepackage{booktabs} 
\usepackage{hyperref}

\begin{document}

\maketitle
\thispagestyle{empty}
\pagestyle{empty}

\begin{abstract}
Quadrupedal robots excel in mobility, navigating complex terrains with agility. However, their complex control systems present challenges that are still far from being fully addressed.
In this paper, we introduce the use of Sample-Based Stochastic control strategies for quadrupedal robots, as an alternative to traditional optimal control laws. We show that Sample-Based Stochastic methods, supported by GPU acceleration, can be effectively applied to real quadruped robots. In particular, in this work, we focus on achieving gait frequency adaptation, a notable challenge in quadrupedal locomotion for gradient-based methods. 
To validate the effectiveness of Sample-Based Stochastic controllers we test two distinct approaches for quadrupedal robots and compare them against a conventional gradient-based Model Predictive Control system. Our findings, validated both in simulation and on a real 21Kg Aliengo quadruped, demonstrate that our method is on par with a traditional Model Predictive Control strategy when the robot is subject to zero or moderate disturbance, while it surpasses gradient-based methods in handling sustained external disturbances, thanks to the straightforward gait adaptation strategy that is possible to achieve within their formulation.

\end{abstract}

\section{INTRODUCTION}

The superior mobility of quadrupedal robots allows them to navigate through uneven landscapes, climb obstacles, and maintain stability on slippery surfaces, showcasing a level of adaptability that closely mirrors the capabilities of their biological counterparts \cite{Biswal2017}. However, the advantages offered by quadrupedal robots come at the price of increased complexity in designing and implementing effective locomotion controllers.
In recent years, the field of robotics has seen a significant surge in the development of innovative control solutions for legged robots, particularly quadrupedal systems \cite{lunardi2024}. This advancement is fueled by the aim to realize sophisticated locomotion strategies that are capable of seamlessly adapting to diverse and challenging real-world scenarios. 

The quest for advanced quadrupedal mobility has pushed researchers to explore a variety of control methodologies, from analytical optimization and Model Predictive Control (MPC) \cite{dicarlo2018} and purely end-to-end Reinforcement Learning strategy \cite{Rudin2022}   to cutting-edge machine learning strategies that blend reinforcement learning with traditional optimization techniques \cite{omar2023}.

\begin{figure}[!t]
\vspace{15pt}
\centering{
\includegraphics[width=0.48\textwidth]{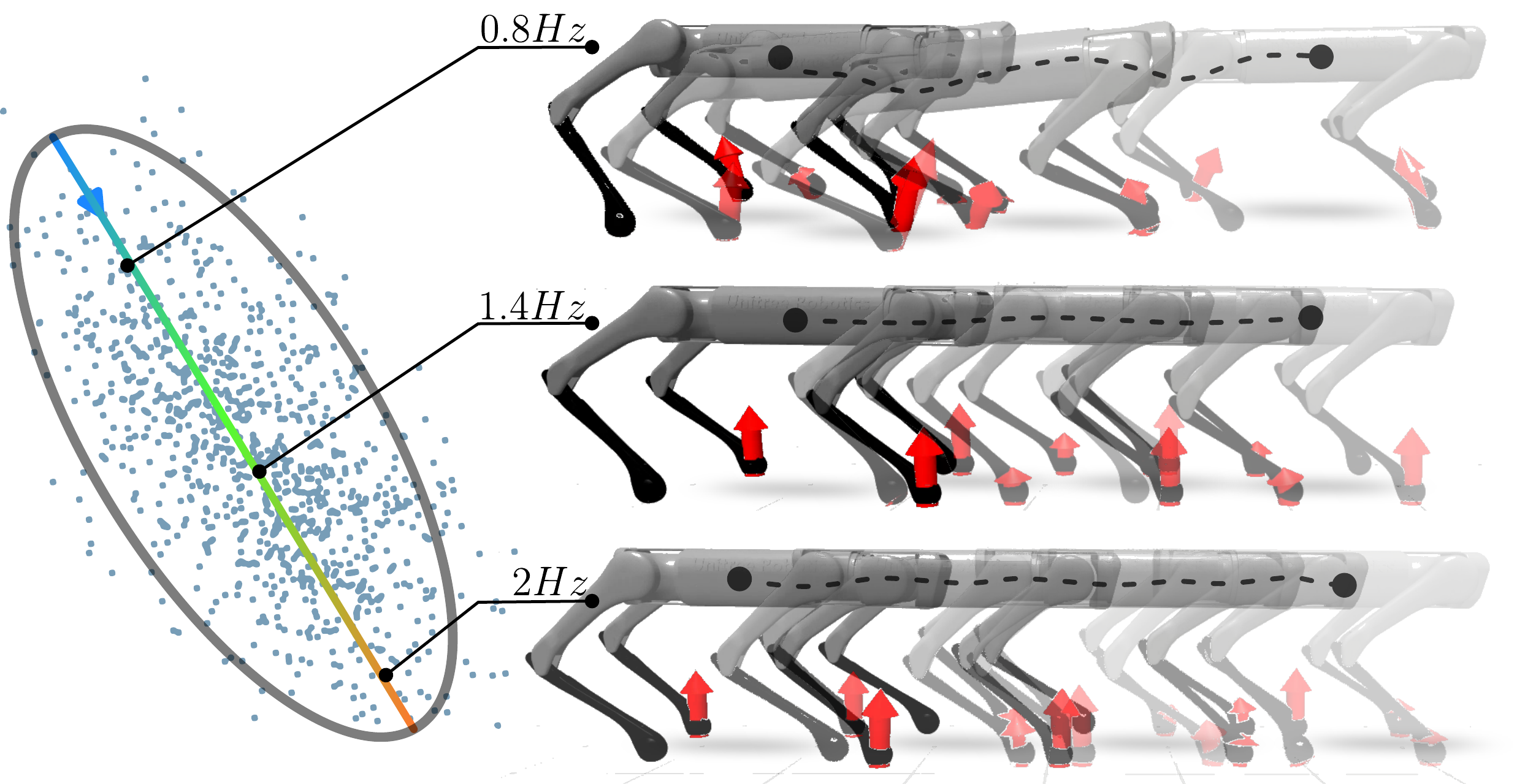}
}    
\caption{The illustration depicts the application of a Sample-Based Stochastic (SBS) predictive controller for gait generation in a quadrupedal robot. Each gait sequence, sampled from a distribution, corresponds to a unique set of control actions at various gait frequencies. These sequences are evaluated based on their predicted cost over the forecast horizon. The most promising sequences—referred to as the elite samples—are then used to update the sampling distribution, optimizing the controller's future action selections.
}
\label{fig:robot_real}
\vspace{-10pt}
\end{figure}
Sample-Based Stochastic (SBS) methodologies are rapidly gaining traction \cite{howell2022} and represent a valid alternative to control mobile robots. This strategy diverges from traditional control methods by relying on a distribution-based generation and evaluation of a multitude of potential solutions sampled from the search space, rather than exclusively depending on gradient-based optimization techniques.

Historically, SBS optimization methods \cite{Stulp2012} have emerged as a compelling alternative to analytical approaches in instances where gradients are intractable or nonexistent. These methods have proven particularly useful in robotic applications where the complexity of the robot's interactions with its environment precludes the straightforward application of gradient-based techniques. SBS methods can uncover effective strategies for controlling robots in diverse and challenging situations by sampling a wide range of possible solutions and evaluating their performance.
We believe that adopting SBS control strategies in quadrupedal robotics could represent a significant advancement, offering a flexible and robust framework for developing locomotion and maneuvering capabilities that can adapt in real-time to the ever-changing demands of real-world environments.

Moreover, the widespread adoption of programming languages and libraries like JAX \cite{Bradbury2018} opens up unprecedented opportunities for implementing real-time SBS control solutions in real robotics platforms. JAX is a high-performance library designed for numerical computing, particularly well-suited for the demands of parallel large-scale computation.

In this study, we show the application of GPU-accelerated, SBS real-time controllers for quadruped robots, demonstrating that SBS methods can match the performance of traditional gradient-based controls while reducing controller complexity. Furthermore, we highlight the SBS controller's adaptability by easily addressing challenges like gait frequency adaptation, a persistent issue in quadrupedal locomotion research \cite{amatucci}.

To summarize, our contributions are:
\begin{itemize}
    \item 
 The first implementation of a GPU-accelerated Sample-Based Stochastic (SBS) control strategy for the direct control of a 12 degrees of freedom quadrupedal robot, fully integrated within a comprehensive locomotion control framework.
    \item Introduction of a straightforward gait adaptation strategy, illustrating the SBS's ability to optimize robot locomotion under external disturbances.
    \item A detailed comparative analysis of two SBS control strategies against established gradient-based optimal controllers, underscoring the practical advantages and effectiveness of the SBS approach in real-world scenarios.
    \item An open-source code repository\footnote{code repository: https://github.com/iit-DLSLab/Quadruped-PyMPC}, for the rapid implementation and testing of SBS control strategies for quadrupedal robots.
\end{itemize}
The paper is organized as follows: Section~\ref{sec:rel_works} reviews related works, providing a foundation and context for our research. Section~\ref{sec:SBS} delves into SBS control methods and their application to the predictive case. In Section~\ref{sec:method}, we explore how SBS predictive control methods can be effectively applied for quadrupedal-legged locomotion. Section~\ref{sec:results} presents the experimental results, showcasing SBS controllers on a real Unitree Aliengo\footnote{aliengo quadruped robot: https://www.unitree.com/products/aliengo}. Finally, Section~\ref{sec:conclusion} concludes the paper, summarizing our findings and suggesting directions for future research.

\section{RELATED WORKS}\label{sec:rel_works}

The challenge of achieving effective quadrupedal locomotion has traditionally been addressed through analytical gradient-based methods. For instance, in the approach outlined by \cite{dicarlo2018}, the authors develop an optimal control strategy based on linearized rigid body dynamics. This strategy demonstrates the potential for practical application in real quadrupedal robots. In \cite{kang2022} the authors employ a Nonlinear Model Predictive Control (NMPC) to track retargeted animal motions, while in \cite{Ruben2023} the authors propose an NMPC approach to traverse rough terrain using elevation maps input to optimize footstep locations.

It is important to note that all these analytical methods often depend on complex structures that require the presence of external solvers, such as qpOASES~\cite{qpOASES}, or HPIPM~\cite{HPIPM}, and gradient information. Even though all these methods display great robustness derived from the inherent stability of the optimal control framework, this strength is somewhat offset by their lack of adaptability, stemming from the necessity for continuous and differentiable costs within their formulations. 

Another promising avenue for addressing locomotion challenges has been  Reinforcement Learning (RL) strategies, which have yielded impressive outcomes in various applications thanks to their ability to handle sparse rewards or costs that are not differentiable. In \cite{cheng2023} and \cite{zhuang2023} the authors propose different RL policies for performing parkour-like motions with real robotic quadrupeds. In \cite{Rudin2022} the authors introduce an end-to-end RL policy for quadrupeds navigation.
Despite the great results achieved by RL approaches, a significant limitation is their lack of generalizability to different tasks without undergoing comprehensive retraining. This constraint means that for each new problem, the RL model often requires a fresh training cycle, which can be resource-intensive. Furthermore, the black-box nature of these methods often presents challenges related to the safe deployment of the policies in the real world.

SBS methods emerge as a powerful hybrid, merging the precision of model-based analytical optimal control techniques with the inherent adaptability of RL approaches. These methods have shown remarkable potential in generating complex behaviors through a limited number of concurrent samples. For instance, in \cite{howell2022, asmar2022}, the efficacy of SBS methods in creating sophisticated behaviors is demonstrated only within simulated environments. Several studies have validated the effectiveness of SBS methods on actual robotic platforms \cite{Pravitra2021, Williams2017}, highlighting their practical applicability. In \cite{Yang2019}, the authors attempted to integrate a sample-based controller with RL, but due to the platform's simplicity and its limited degrees of freedom, the application fell short of a comprehensive locomotion scenario. Despite a wide array of methods being successfully applied across various robotic systems, there remains a noticeable gap in their application to locomotion. Finally in \cite{hutter_sampling}, the authors introduce an SBS strategy with an auxiliary policy for real-time operation due to CPU limitations on sample size. Conversely, our GPU-accelerated SBS controller allows for direct, real-time robot control without such limitations.

\section{SAMPLE-BASED STOCHASTIC OPTIMIZATION}\label{sec:SBS}

SBS methods, which have been utilized extensively in the last decade in robotics applications \cite{Modugno2017, Stulp2012}, often employ the multivariate Gaussian distribution as the cornerstone for exploring the parameters landscape. This distribution, denoted as $\mathcal{N}(\boldsymbol{\theta}, \Cm)$, where $\boldsymbol{\theta}$ represents the mean and $\Cm$ the covariance matrix, provides a probabilistic framework for generating parameter samples. Within this context, it is possible to identify a general structure for SBS optimization as shown in Algorithm~\ref{alg:SO}. 

\begin{algorithm}
\caption{Generic iteration of Sample-Based Stochastic optimizer}\label{alg:SO}
\label{alg:generic_iteration_SBS}
\begin{algorithmic}[1]
\State $\thetav_{k=1 \ldots K} \sim \mathcal{N}(\thetav, \Cm)$ \Comment{\textbf{Sampling}}
\State $J_k = \text{CostEval}(\thetav_k)$ \Comment{\textbf{Evaluation}}
\State $\thetav_{k=1 \ldots K}, \leftarrow \text{sort}(\thetav_{k=1 \ldots K}, J_{k=1 \ldots K})$ \Comment{\textbf{Sorting}}
\State $J_{k=1 \ldots K} \leftarrow \text{extract sorted costs}$
\State \textbf{Update}:
\begin{itemize}
    \item $\thetav^{\text{new}} = \text{UpdateMean}(\thetav_{k=1 \ldots K_e},J_{k=1 \ldots K_e})$ 
    \item $\Cm^{\text{new}} = \text{UpdateCov}(\thetav_{k=1 \ldots K_e}, J_{k=1 \ldots K_e})$ 
\end{itemize}
\end{algorithmic}
\end{algorithm}

The process starts with the \textbf{sampling} phase, where $K$ samples $\boldsymbol{\theta}_{k=1 \ldots K}$ are drawn from the multivariate Gaussian distribution. Following the sample generation, an \textbf{evaluation} and \textbf{ordering} steps are performed, where each parameter set $\boldsymbol{\theta}_k$ is evaluated using the cost function $J\left(\boldsymbol{\theta}_k\right)$. The samples are then ranked in ascending order of their cost values, prioritizing those that indicate lower costs. Finally, in the \textbf{update} phase, the algorithm refines the parameters of the Gaussian distribution, concentrating on the 'elite' samples. This subset consists of the top $K_e$ samples from the sorted list, representing the most promising directions for further exploration. The mean and covariance are updated to guide the subsequent sampling toward more promising regions of the parameter space. To converge to the optimal solution of the cost function, it is necessary to execute these four phases multiple times iteratively.

\subsection{Sample-Based Stochastic Methods for Predictive Control}

The essence of applying SBS methods lies in performing a complete rollout over a certain horizon to compute the cost function to optimize w.r.t. to the decision variables.
Therefore in the context of predictive SBS with a horizon of $N$ steps, the $\text{CostEval}(\cdot)$ function defined for computing $J_k$ associated with the kth sample becomes:
\begin{algorithm}
\caption{Rollout($\thetav_k$)}\label{alg:rollout}
\begin{algorithmic}[1]
\For{$i = 0$ to $N-1$}
    \State $\uv_i = \piv(\thetav_k, \xv_i,t_i)$
    \State $\xv_{i+1} = \fv_i(\xv_i,\uv_i)$
    \State $J_k = J_k + r(\uv_i,\xv_i,\xv^{r}_i)$
\EndFor
\State $\text{return} \, J_k$
\end{algorithmic}
\end{algorithm}

where $\piv(\thetav_k, \xv_i, t_i)$ represents a control policy that depends on the sample decision variables $\thetav_k$, $\fv_i(\xv_i,\uv_i)$ represents the discretized dynamics of the system to be controlled, $r(\uv_i,\xv_i,\xv^{r}_i)$ represents the one-step cost associated with the current state-input pair, and $\xv^{r}_i$ defines the desired state reference.

Using sampling-based methods in the context of MPC is simple in practice because it does not require the use of any optimizers, like HPIPM or qpOASES. Classical optimizers often come with a host of practical challenges such as scalability issues, handling discontinuities, and dependency on gradient information. 
On the other hand, SBS methods, being gradient-free, offer the flexibility to compute solutions across a wide range of scenarios, including those with discontinuities, which are particularly prevalent in multi-contact systems. 

Even though SBS methods are traditionally viewed as sample-inefficient, 
this issue is greatly mitigated by leveraging the computational power of modern GPUs. 
This parallelization does not only enhance efficiency, but also allows for the randomization of other features of the control problem. Such capabilities significantly expand the practical applicability of SBS methods in MPC, as we will illustrate in the subsequent section in the context of legged locomotion.

In a predictive control scenario, for the sake of efficiency, we conduct just a single iteration of Algorithm~\ref{alg:generic_iteration_SBS} at every time step. To enhance the search process for successive iterations, we employ a strategy known as \textit{warm-starting}. This entails initiating each new search from the solution obtained in the previous iteration. This significantly accelerates the optimization cycle, ensuring that our methods remain computationally viable even in scenarios demanding real-time decision-making. 

In this work, we employed two SBS predictive methods: the Na\"{i}ve SBS optimizer and the Model Predictive Path integral (MPPI) optimizer \cite{Theodorou2010}.

\subsection{Na\"{i}ve SBS Optimizer}
We introduced the Na\"{i}ve SBS method as a baseline to demonstrate the effectiveness of SBS methods, even when the algorithm's structure is markedly simple. In this approach, we employ a straightforward  UpdateMean($\cdot$) strategy where, at each iteration, the chosen elite sample is always the best-performing one. To maintain the algorithm's simplicity, we do not update the covariance from one iteration to the next, leaving it unchanged. This decision to keep the covariance constant is crucial for sustaining robust exploration capabilities, which we find to be highly beneficial in this context.

A generic iteration of the Na\"{i}ve SBS Optimizer is summarized in Algorithm~\ref{alg:SO_naive}, where with $\thetav^{*}$ we define the best sample drawn during one iteration. 
\begin{algorithm}
\caption{One iteration of the Na\"{i}ve Optimizer}\label{alg:SO_naive}
\begin{algorithmic}[1]
\State $\thetav_{k=1 \ldots K} \sim \mathcal{N}(\thetav, \Cm)$ 
\State $J_k = \text{Rollout}(\thetav_k)$ 
\State $\thetav_{k=1 \ldots K} \leftarrow \text{sort}(\thetav_{k=1 \ldots K}, J_{k=1 \ldots K})$ 
\State \textbf{Update}:
\begin{itemize}
    \item $\thetav^{\text{new}} = \thetav^{*}$ \Comment{Current best as new mean}
    \item $\Cm^{\text{new}} = \Cm$ \Comment{Covariance is unchanged}
\end{itemize}
\end{algorithmic}
\end{algorithm}

\subsection{Model Predictive Path Integral Optimizer}
The MPPI algorithm originates from fundamental optimal control principles and derives its name from utilizing the Feynman-Kac lemma. This approach reformulates the Hamilton-Jacobi-Bellman (HJB) equations into evaluating an expectation over all possible trajectories (or paths) that the system could take, weighted by their performance of path integral that can be effectively estimated using Monte Carlo methods as shown in \cite{Theodorou2010}. 
This weighting scheme reflects the principle that trajectories leading to lower costs are more likely to be close to the optimal path.
The rollout weighting procedure is defined as
\begin{align*}
    &\tilde{\omega}_i = \exp\left(-\frac{1}{\lambda} \cdot (J_{i} - \beta)\right) 
(\omega_i \cdot \thetav_{i})
\end{align*}
\begin{align*}
    &\Omega = \sum_{i=1}^{K} \tilde{\omega}_i   
    &\thetav^{\text{new}} = \sum_{i=1}^{K}(\omega_i \cdot \thetav_{i})
\end{align*}
where $\beta = J_1, \quad \lambda = 1 $ represent respectively a normalization factor and the temperature parameter of MPPI, $ \tilde{\omega_i}$ and $\omega_i=\frac{\tilde{\omega}_i}{\Omega}$ represent the weight and the normalized weight associated with each sample $\thetav_i$. A full representation of the MPPI algorithm is shown in Algorithm~\ref{alg:SO_MPPI}.
\begin{algorithm}
\caption{One iteration of the MPPI Optimzer}\label{alg:SO_MPPI}
\begin{algorithmic}[1]
\State $\thetav_{k=1 \ldots K} \sim \mathcal{N}(\thetav, \Cm)$ \Comment{Sampling}
\State $J_k = \text{Rollout}(\thetav_k)$ \Comment{Evaluation}
\State $\thetav_{k=1 \ldots K} \leftarrow \text{sort}(\thetav_{k=1 \ldots K}, J_{k=1 \ldots K})$ \Comment{Sorting}
\State $J_{k=1 \ldots K} \leftarrow \text{extract sorted costs}$
\State \textbf{Update:}
\begin{itemize}
    \item  $\text{UpdateMean}(\thetav_{k=1 \ldots K}, J_{k=1 \ldots K})$
    \item $\Cm^{\text{new}} = \Cm  $ \Comment{Covariance is unchanged}
\end{itemize}

\item[]
\Function{$\text{UpdateMean}$}{$\thetav_{k=1 \ldots K}, J_{k=1 \ldots K}$}
\State $\beta = J_1$ \Comment{Best cost after sorting}
\State $\lambda = 1$
\State Initialize $\tilde{\omega}_i$ for each sample
\For{$i = 1$ to $K$}
    \State $\tilde{\omega}_i = \exp\left(-\frac{1}{\lambda} \cdot (J_{k_i} - \beta)\right)$ 
\EndFor
\State $\Omega = \sum_{i=1}^K \tilde{\omega}_i$
\For{$i = 1$ to $K$}
    \State $\omega_i = \frac{\tilde{\omega}_i}{\Omega}$ \Comment{Normalize weights}
\EndFor
\State $\omega_{k=1\ldots,K} \leftarrow [\omega_1,\ldots,\omega_K]$
\State $\thetav^{\text{new}} = \sum_{i=1}^K(\omega_{k=1\ldots,K} \cdot \thetav_{k=1 \ldots K})$
\EndFunction

\end{algorithmic}
\end{algorithm}

\section{LEGGED LOCOMOTION WITH SBS PREDICTIVE CONTROLLER}\label{sec:method}
In this section, we demonstrate the applicability of the SBS predictive control framework to the problem of legged locomotion. Specifically, we showcase the versatility of SBS methods in extending beyond basic locomotion control to optimize additional aspects of robot walking, such as gait frequency, which is known to be complex to optimize within a gradient-based approach in real-time~\cite{winkler}. 

Our proposed control architecture operates by sampling $K$ sets of decision variables, denoted as $\thetav = [\thetav_1, \thetav_2]$, at each time step. Here, $\thetav_1$ controls the method $\text{computeContactSequence}(\thetav_1)$ which generates different gaits $\deltav$, whereas $\thetav_2$ determines the behavior of the mapping policy $\Gammam=\sigmav(\thetav_2)$ that provides the input contact forces $\Gammam$. To evaluate each parameter set $\thetav_k$, we calculate the associated costs performing parallel rollout on the GPU. For this, we employ a discretized model $\fv(\xv,\uv,\deltav)$ based upon the Single Rigid Body Dynamics (SRBD) to advance the system dynamics over $N$ steps and a single step cost $r(\uv_i,\xv_i,\xv_i^{r})$, which represent the tracking cost between the reference state $\xv_i^{r}$ and the current system state $\xv_i$. Following this evaluation, we update the mean of the Gaussian distribution $\mathcal{N}(\thetav, \Cm)$, which is then passed to a low-level control block that computes the robot's torques. In parallel, we compute the foothold reference where the robot needs to step in, which is then passed to the next MPC control loop, and the swing trajectory for the legs that are not currently in contact with the ground. This separation of tasks is generally adopted for controllers that utilize the SRBD model. More details are provided in Section~\ref{subsec:model}. 

\begin{figure*}[!t]
\vspace{-5pt}
\centering{
\includegraphics[width=0.93\textwidth]{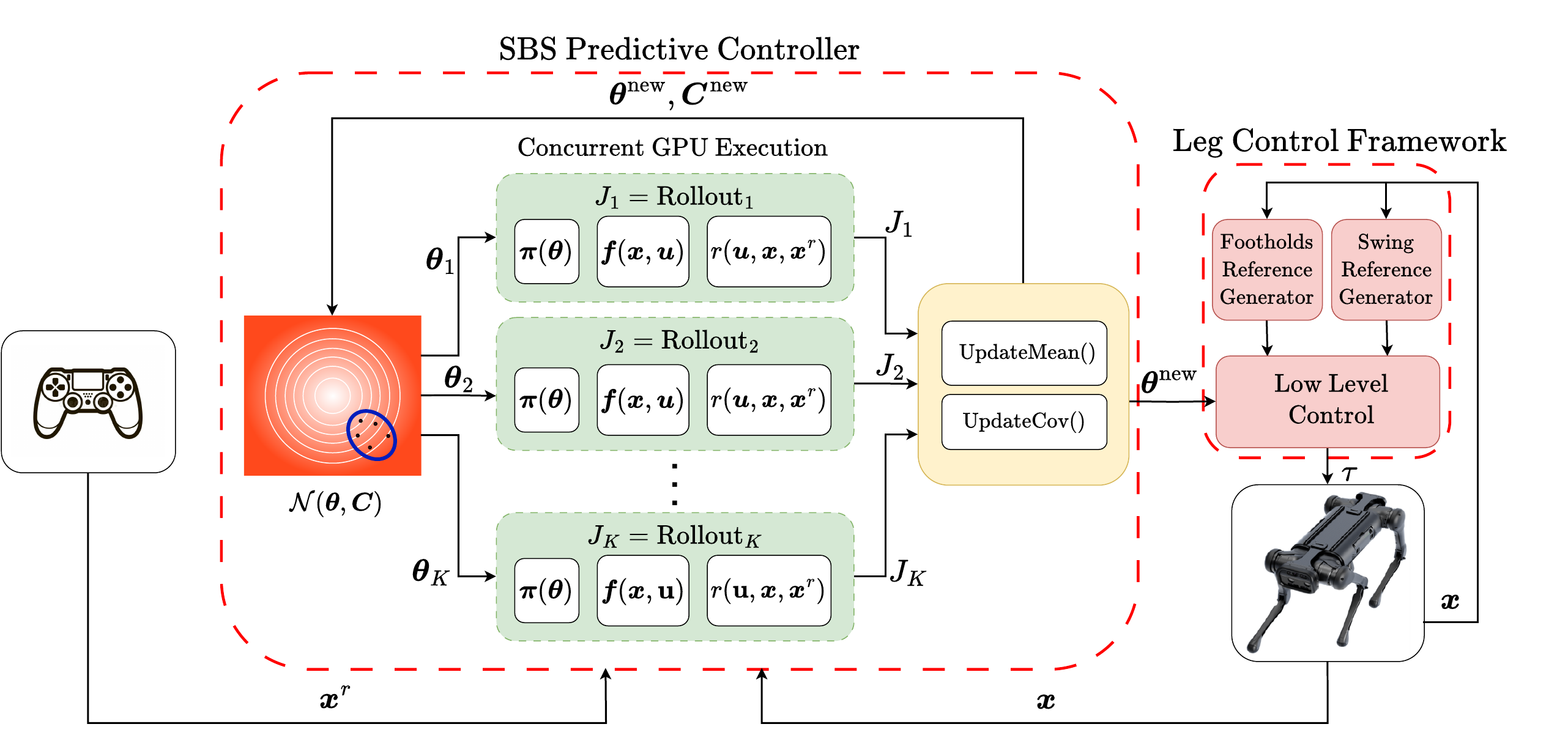}
}    
\caption{Block scheme of the proposed control method. From left to right: first, we command the robot with a user-defined reference, which is then passed to the SBS controller that performs a parallel evaluation of the best control actions in GPU; concurrently, a leg control framework generates foothold references, swing trajectories, and low-level control actions, considering the MPC solution.}
\label{fig:block_scheme}
\vspace{-10pt}
\end{figure*}

A depiction of the methodology is presented in the blocks scheme in Figure~\ref{fig:block_scheme}, where with \textit{Leg Control Framework} we refer to the foothold and swing references, and the low-level control computation routines described above. Finally, a detailed description of the SBS predictive controller adapted for legged locomotion is provided in Algorithm~\ref{alg:Pipeline}.


\begin{algorithm}[h!]
\caption{SBS Predictive Controller for Locomotion}
\begin{algorithmic}
    \State $\textbf{Given:}$
    \State $\xv_{0} \, \text{current state estimation}$
    \State $\thetav_{k=1 \ldots K} \sim \mathcal{N}(\thetav, \Cm)$ 
    \For{$\textbf{each sample } k \text{ (parallel GPU execution)}$}
    \State $J_k = \text{Rollout}(\thetav_k,\xv_{0})$  \Comment{See Algorithm~\ref{alg:rollout} }
    \EndFor
     \State $\thetav_{k=1 \ldots K} \leftarrow \text{sort}(\thetav_{k=1 \ldots K}, J_{k=1 \ldots K})$ 
     \item $[\thetav^{\text{new}}_1,\thetav^{\text{new}}_2] = \text{UpdateMean}(\thetav_{k=1 \ldots K_e},J_{k=1 \ldots K_e})$ 
    \item $\Cm^{\text{new}} = \Cm$

\item[]
\Function{$\piv$}{$\thetav, t_j$}
\State $[\thetav_1,\thetav_2] \gets \thetav$
\State $\deltav_j \gets \text{computeContactSequence}(\thetav_1)$
\State $\Gammam_j \gets \sigmav((t^\text{knot}_{1,\ldots,P}, \thetav_{2,k}),  t_j)$
\State $\uv_j=[\Gammam_j,\deltav_j]$
\EndFunction

\end{algorithmic}
\label{alg:Pipeline}
\end{algorithm}

\subsection{Model Formulation}\label{subsec:model}

This section outlines the model-based approach for a quadrupedal robot used in the $\text{Rollout}(\cdot)$ function, adopting a simplified Single Rigid Body Dynamics (SRBD) model from \cite{Rathod2021}. This model focuses on the quadruped's core translational and rotational movements, omitting the swinging legs' dynamics. This is a suitable approximation since the bulk of a quadruped's mass is typically in its trunk. The robot's dynamics are described using two reference frames: an inertial frame $\mathcal{W}$, and a body-aligned frame $\mathcal{C}$ at the Center of Mass (CoM), simplifying the inertia tensor representation. The SRBD model is articulated through a state representation in the CoM frame

\begin{equation}  
\hspace{-5pt}
\resizebox{\columnwidth}{!}{$\left[\begin{array}{c}\dot{\pv}_{\mathrm{c}} \\ \dot{\vv}_{\mathrm{c}} \\ \dot{\Phi} \\ \dot{\mathcal{\omegav}}\end{array}\right]=\left[\begin{array}{c}\vv_{\mathrm{c}} \\ 1 / m \sum_{i=1}^4 \delta_i \Gammam_i+\gv \\ \Em^{\prime-1}(\Phim)_{\mathcal{C}} \omegav \\ -{ }_{\mathcal{C}} \IIm_{\mathrm{c}}^{-1}\left({ }_c \omegav \times{ }_{\mathcal{C}} \IIm_{\mathrm{c}}\right)_{\mathcal{C}} \omegav+\sum_{i=1}^4 \delta_{i \mathcal{C}} \IIm_{\mathrm{c}}^{-1}{ }_{\mathcal{C}} \pv_{\mathrm{cf}, i} \times{ }_{\mathcal{C}} \Gammam_i\end{array}\right]$}
\label{eq:srbd_model}
\end{equation}

with $m$ characterizing the robot mass subjected to gravitational acceleration $\gv$. Moreover, $\vv_c \in \mathbb{R}^3$ and $\dot{\vv}_c \in \mathbb{R}^3$, respectively, depict the CoM velocity and acceleration. The interaction between the robot and the environment is mediated by the Ground Reaction Forces (GRFs) $\Gammam_i \in \mathbb{R}^3$ at each foot $i$, while the robot's orientation and motion dynamics are captured by the inertia tensor ${}_{\mathcal{C}}\IIm_c \in \mathbb{R}^{3 \times 3}$ at the CoM and the base's angular acceleration ${}_{\mathcal{C}}\dot{\omegav} \in \mathbb{R}^3$. The positional relationship between the CoM $\pv_c \in \mathbb{R}^3$ and each foot $i$'s position $\pv_{f,i} \in \mathbb{R}^3$ is denoted as $\pv_{cf,i} \in \mathbb{R}^3$.  $\Phi = (\phi, \theta, \psi)$ represents the robot body orientation where $\phi, \theta, \psi,$ are the roll, pitch, and yaw respectively, while $\Em^{\prime-1}$ is a mapping from the SRBD angles to Euler rates (see \cite{Rathod2021}).

To model the capacity of each foot $i$ to exert contact forces, binary parameters $\delta_i = \{0, 1\}$ are employed, indicating the presence or absence of ground contact where the index $i$ indicates one of the four legs. The system's state and control are represented by vectors $\xv = (\pv_c, \vv_c, \Phi, {}_{\mathcal{C}}\omegav)$ and $\Gammam= (\Gammam_1, \ldots, \Gammam_4)$. The input to the system model can be defined as $\uv=[\Gammam,\deltav]$, making it possible to formulate the discrete dynamics of the robot as:
\[ \xv_{j+1} = \fv_i(\xv_j, \uv_j), \]
where $\boldsymbol{\delta}_j$ encapsulates the contact state of all four legs at the $j$th timestep, which normally, in model-based controllers, are pre-computed to attain a tractable optimization problem (see Section~\ref{subsec:gait_opt}).

In the following, we show how we parametrize the GRFs and the contact parameters $\deltav_j$ to solve the aforementioned control problem efficiently.

\subsection{GRFs Parametrization}\label{subsec:grf}

In this study, we implement a simple parametrization to represent the actual control actions $\Gammam_i$ for input into the SRBD model over the prediction horizon $N$. The introduction of this intermediate representation serves a dual purpose: firstly, it enables the projection of the control variable across the prediction horizon into a more manageable, lower-dimensional space; secondly, it inherently ensures the smoothness of the control actions. As in \cite{howell2022}, our approach utilizes cubic splines for this task. Specifically, within each sample $\thetav_k$, the decision variable $\thetav_{2,k} \in \mathbb{R}^{P}$ represents a time-indexed sequence of $P$ knots, each allocated to specific time instants $t^{\text{knot}}_{1,\ldots,P}$. 

For any given input time $t$, the spline's evaluation is then determined as
\begin{equation}
\Gammam = \sigmav((t^\text{knot}_{1,\ldots,P}, \thetav_{2,k}),  t)
\end{equation}

and its output value is then constrained to respect friction cone constraints to guarantee non-slipping conditions.

\subsection{Gait Optimization}\label{subsec:gait_opt}

Gradient-based methods for legged locomotion often precompute the contact status of each leg as a fixed vector $\deltav$ due to the challenges posed by Linear Complementary Constraints, which complicate analytical gradient computations~\cite{NMPC_contact}. To address this, fixed periodic gaits like trotting or pacing are commonly used for their simplicity and effectiveness, despite the trade-off between robustness and energy efficiency~\cite{different_gait}. While some studies have attempted to optimize gait periodicity using heuristics, these optimizations typically occur outside the control loop~\cite{amatucci}, \cite{Boussema}.

SBS controllers, however, are not limited by the constraints of gradient-based methods, allowing for the direct optimization of gait parameters within the MPC framework. A periodic gait is characterized by the step frequency $f_s$ and the duty factor $D_f$, defining the stance and swing times for each leg as $T_{st} = \frac{D_f}{f_s}$ and $T_{sw} = \frac{1 - D_f}{f_s}$, respectively.

This work optimizes step frequency $f_s$ by discretizing it over a fixed range and parametrizing it with $\thetav_1$. The function $\text{computeContactSequence}(\thetav_1)$ calculates lift-off events and $\deltav$ for the MPC horizon. To promote energy efficiency, we introduce a regularization term in $J$ for $\thetav_1$ towards a nominal stepping frequency.

\subsection{Leg Control Framework}

The outer control loop, operating concurrently with the MPC, consists of three modules: a \textit{foothold reference generator}, a  \textit{swing reference generator}, and a  \textit{low-level} torque command module.

The foothold reference generator calculates the stepping points for each leg using:

\begin{equation}
\pv_{f,i} = \pv_{hip,i} + \frac{T_{st}}{2}\vv_c^d + \sqrt{\frac{\pv_{c,z}}{g}}(\vv_c - \vv_c^d),
\label{eq:foothold}
\end{equation}

where $\pv_{hip,i}$ represents the hip position, $\vv_c^d$ the desired velocity, and $\pv_{c,z}$ the height of the CoM. The last term in (\ref{eq:foothold}) adjusts footholds for handling external disturbances.

Swing trajectories for legs in the air are generated by the swing reference generator using cubic splines, linking lift-off points to the touchdown points $\pv_f$ with a nominal stepping height.

The low-level module translates Ground Reaction Forces from Algorithm~\ref{alg:Pipeline} into motor torques for stance legs as $ \tauv_{st} = -\Jm^\top(\qv)\Gammam$, and for swing legs using inertia, Jacobian matrices, and PD control for foot trajectory adherence.

These components ensure dynamic, disturbance-compensated locomotion with precise control over foot placement and swing motion.

\section{RESULTS}\label{sec:results}

We validated our proposed controllers through simulations and real-world experiments on the 21kg Unitree Aliengo robot. We compared three controllers:

\begin{enumerate}
  \item acados \cite{Verschueren2019}: an optimization-based approach using the same model from Section~\ref{sec:method}-A,
  \item Na\"{i}ve Optimizer: described in Section~\ref{sec:SBS}-B,
  \item MPPI: detailed in Section~\ref{sec:SBS}-C.
\end{enumerate}

Their performance was evaluated under various disturbances in Mujoco simulations~\cite{todorov2012mujoco} and actual hardware tests. All methods were tested with a horizon length of \(N = 12\), a discretization time of 0.02 seconds, and a trotting gait using a nominal stepping frequency of \(f_s^{n}= 1.3\)Hz. Acados ran on an Intel 13700H CPU, while the SBS controllers utilized an Nvidia 4050 mobile GPU that can draw a maximum of $30$W of power, making it ideally suitable for being carried directly on board a quadruped robot.

We used a quadratic cost function for defining the control task
\begin{equation*}
    J = (\xv - \xv^r)^\top \Qm (\xv - \xv^r) + (\uv - \uv^r)^\top \Rm (\uv - \uv^r),
\end{equation*}
where \(\Qm\), \(\Rm\), are positive diagonal weighting matrices.

For gait optimization, we regularized the stepping frequency towards a nominal, more energy-efficient value, using
\begin{equation*}
    J = J + \rho (\thetav_1 - \thetav^r_1)^2,
\end{equation*}
with \(\rho\) as a scalar weight and \(\thetav^r_1 = f_s^{n}\). The variable \(\thetav_1\) ranged between \([f_s^{n}, 2.4]\)Hz, and sampled using a uniform distribution. We choose to discretize for simplicity $\thetav_1$ with three different values, such as $f_s^{n}$, $2.0$, $2.4$, with $f_s^{n}$ defining a nominal more energy-efficient gait while $2.4$ a more robust one, but a more fine-grained discretization step can be employed at need.
\begin{figure}[!t]
\vspace{-15pt}
\centering{
\includegraphics[width=0.48\textwidth]{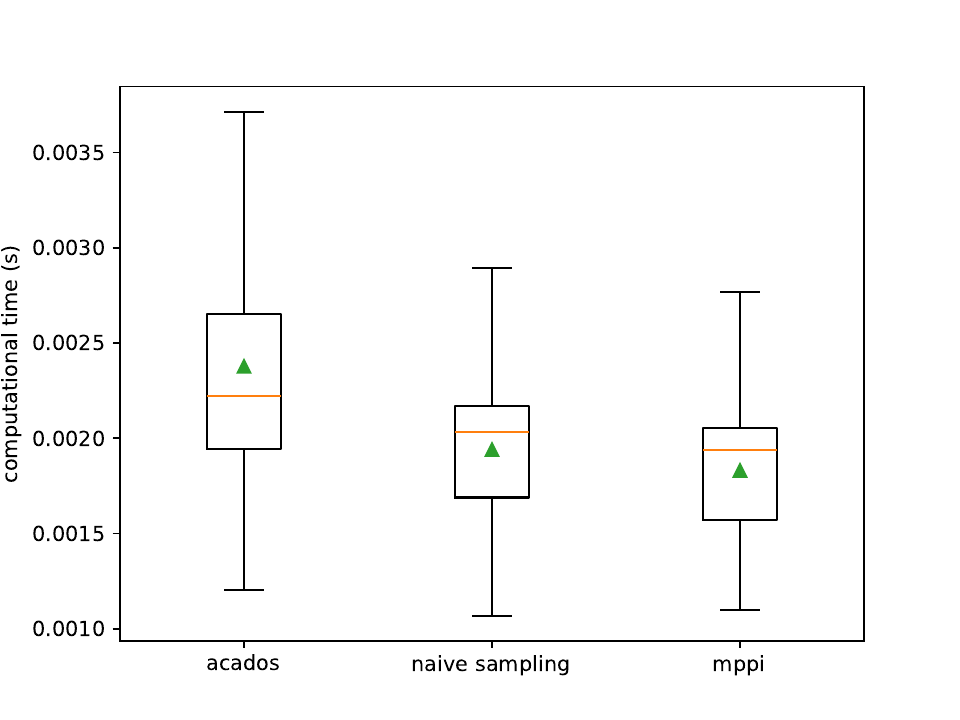}
}    
\vspace{-15pt}
\centering{
\includegraphics[width=0.48\textwidth]{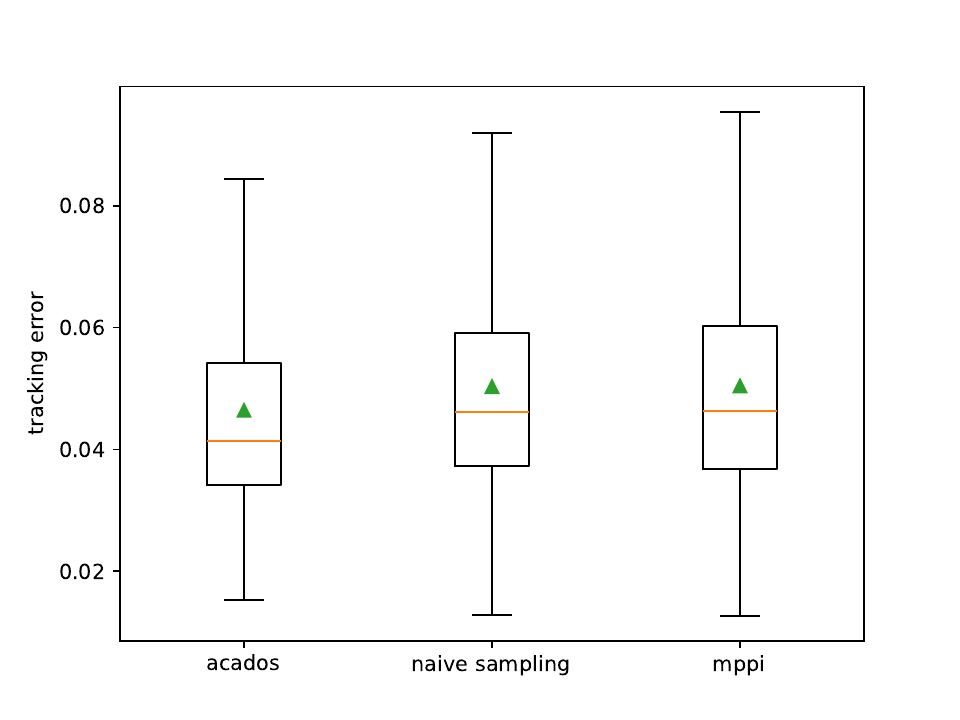}
}
\vspace{-15pt}
\caption{Comparison between acados, na\"{i}ve optimizer, and mppi, commanding to the robot a forward linear velocity of 0.5 m/s. Top: box plot of the computation time (SBS methods perform 10000 rollouts at each time step); bottom: box plot of the tracking error. The box extends from the first to the third quartile of the data, with the orange line and the green triangle that depict the median and the mean values, respectively.
}
\vspace{-5pt}
\label{fig:computational_time}
\end{figure}

\subsection{Simulation Results without Gait Adaptation}

We evaluated the computational efficiency and tracking accuracy of each control method, with SBS methods optimizing over 10,000 rollouts. We aim to demonstrate the practicality of SBS methods, benefiting from simple implementation without dependency on QP solvers, as detailed in Section~\ref{sec:SBS}-A. Simulations involved random external disturbances (wrenches within +/- 5N/Nm) applied to the robot CoM with a duration of $2$ seconds every $2$ seconds.

Figure~\ref{fig:computational_time} shows competitive computational times for all methods and slightly quicker execution for SBS approaches, advantageous for minimizing control delays. In our case, MPPI is faster than (2) since in the last we sample from multiple Gaussian distributions.
Tracking performance under a forward velocity command of 0.5m/s was comparable across methods, with acados (1) marginally outperforming the Na\"{i}ve approach (2) and MPPI (3) as shown in Figure~\ref{fig:computational_time} thanks to the more fine-grained control action obtained by exploiting the gradient information.

During our tests, we have observed that with a large sample size, differences between methods (2) and (3) are less noticeable, but become significant with fewer rollouts. Hence, in the following, we compare acados and Na\"{i}ve sampling for brevity.

\subsection{Simulation Results with Gait Adaptation}

\begin{figure}[!t]
\vspace{-15pt}
\centering{
\includegraphics[width=0.46\textwidth]{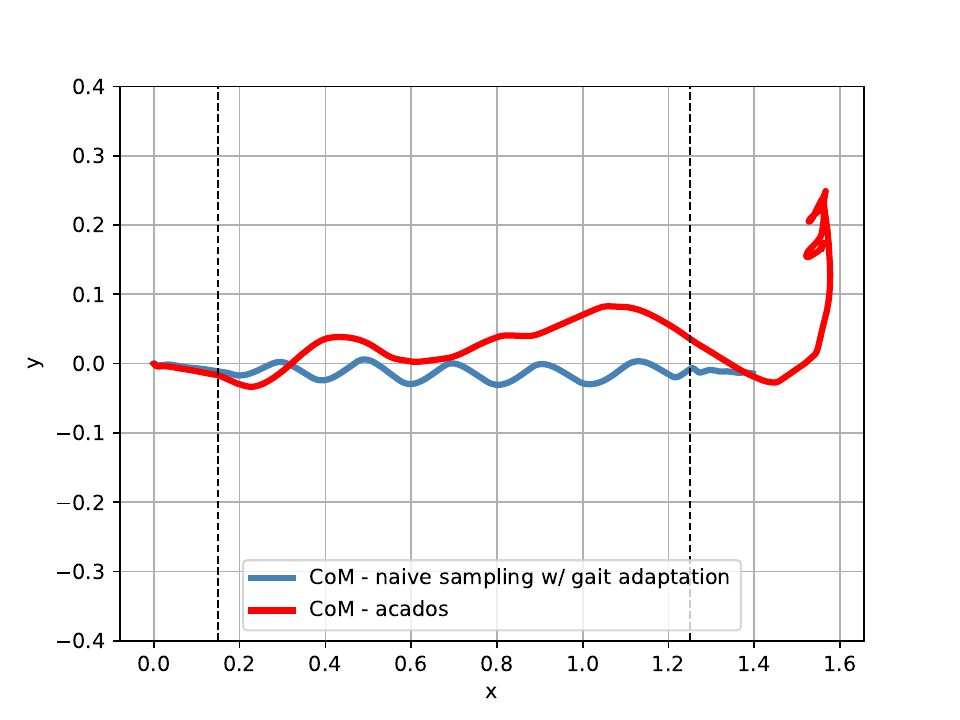}
}   
\includegraphics[width=0.46\textwidth]{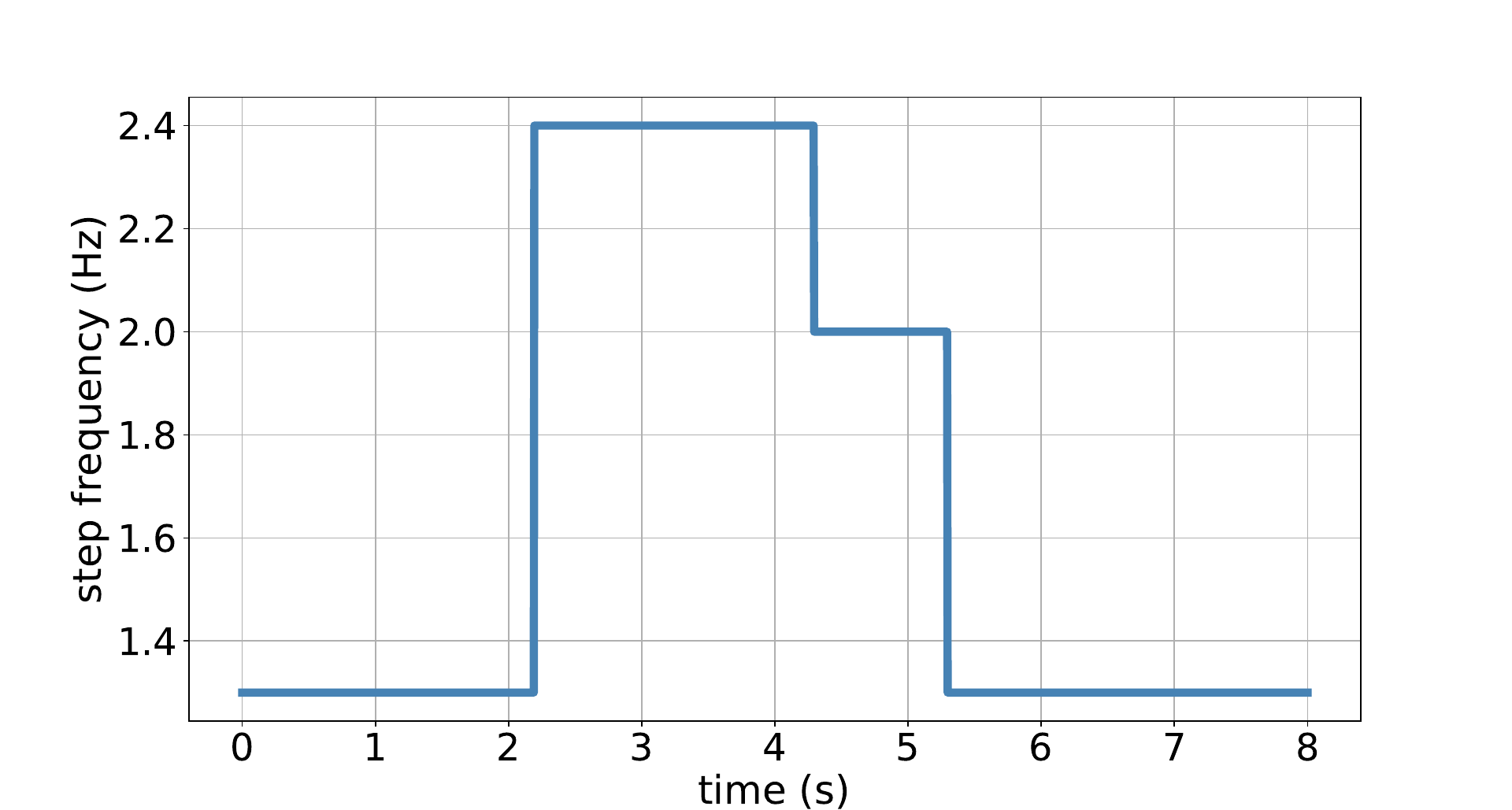}
\vspace{-7pt}
\caption{Comparison between acados and na\"{i}ve optimizer with gait adaptation, under a strong lateral disturbance (40N) applied for $3$ seconds. Top: the CoM evolution under the two control laws, with the dotted line that represents the position range (for the SBS method) in which the external disturbance is applied to the robot; bottom: the step-frequency adaptation profile achieved with the na\"{i}ve optimizer.}
\label{fig:performance_gait_adaptation}
\vspace{-10pt}
\end{figure}

The straightforward simplicity of sampling-based methods and their capability of optimizing over zero-gradient information, as explained in Section~\ref{sec:method}-C, make sampling-based controllers an appealing method for legged locomotion. To validate this claim, in Figure~\ref{fig:performance_gait_adaptation}, we compare the CoM trajectory of Aliengo using acados and na\"{i}ve sampling with gait adaptation using a strong later push of 40N applied at the CoM of the robot. In this simulation, we furthermore commanded a lateral velocity of 0.1 m/s to the system. As can be seen in Figure~\ref{fig:performance_gait_adaptation} (bottom), na\"{i}ve sampling can optimize the stepping frequency at need, increasing the overall robustness of the controller and avoiding the loss of stability of the robot which occurs with the gradient-based controller. In fact, after the application of the disturbance at around second 2, the stepping frequency is automatically increased, and after the disturbance vanishes (around second 4.2), the frequency is slowly restored to the nominal value $f_s^{n}$.

We confirmed this result statistically by analyzing the success rate of methods (1), (2) with and without gait adaptation over 50 different episode simulations, where we applied to the robot CoM random wrench disturbances in the range of +/- 20 [N,Nm]. In Table~\ref{tab:analysis}, we report the obtained results, showing that the gait adaptation strategy embedded inside the SBS controller greatly increases the system's capability to prevent accidental falls and mean tracking errors.
The reader can refer to the accompanying video\footnote{video: https://youtu.be/E4iz9fZsfxA} to visualize the simulated results.

\begin{table}[h!]
    \caption{Statistical Analysis over 50 different episodes}
	\begin{center}
		\begin{tabular}{@{} l l l l l @{}}
			\toprule[0.4mm]
			\textbf{Description} & \textbf{Mean Cost}  & \textbf{Success Rate \%} \\ 
			\midrule
			acados & \hspace{4mm}0.12&  \hspace{7mm}32 \\
			na\"{i}ve sampling& \hspace{4mm}0.14&  \hspace{7mm}37\\
			na\"{i}ve sampling w/ gait adaptation& \hspace{4mm}\textbf{0.08}&  \hspace{7mm}\textbf{80} \\
			\bottomrule[0.4mm]
		\end{tabular}
		\label{tab:analysis}
	\end{center}
 \vspace{-5pt}
\end{table}

\begin{figure*}[!t]

\MujocoAdaptive

\includegraphics[width=1.\textwidth]{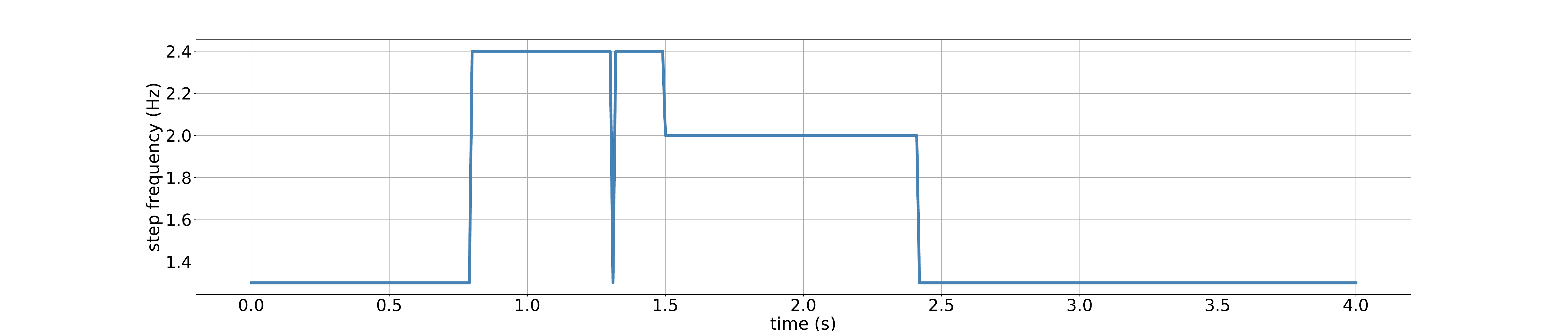}
\caption{Top: snapshots of the robot response during trotting, controlled by the na\"{i}ve sampling technique, to an external disturbance. Bottom: step-frequency adaptation profile during the same experiment.}
\label{fig:acados_vs_sampling_shot}
\vspace{-15pt}
\end{figure*}

\subsection{Experiment Results on Real Quadruped}
The na\"{i}ve sampling controller was validated on the real robot on two different gaits, such as trotting and pacing (see the accompanying video). Furthermore, we replicated the simulation results shown in Figure~\ref{fig:performance_gait_adaptation}, perturbing both acados and the sampling controller with strong lateral pushes. As can be seen in Figure~\ref{fig:acados_vs_sampling_shot}, we obtain a similar behavior to the one achieved in simulation, with the sampling-based controller able to modulate the step frequency at need. In the accompanying video, we report on the same experiment performed using method (1).


\section{CONCLUSION}\label{sec:conclusion}

In conclusion, this paper demonstrated the efficacy of SBS control strategies for quadrupedal robots, offering a robust alternative to conventional gradient-based controls. Our study highlights the SBS methods' success in real-world applications, particularly in achieving real-time gait frequency adaptation—a significant hurdle for gradient-based approaches.

Comparative testing against traditional MPC approaches underline the SBS controllers' comparable performance under minimal or moderate disturbances, and superior handling of severe disturbances due to their flexible formulation that allows the optimization of other controller aspects like the gait frequency.

Future work will focus on applying SBS controllers with the full dynamics of quadrupedal systems for direct torque control and integrating visual feedback to enhance their operational capabilities in complex environments.






\bibliographystyle{IEEEtran}
\bibliography{biblio}

\begin{thebibliography}{10}
\providecommand{\url}[1]{#1}
\csname url@rmstyle\endcsname
\providecommand{\newblock}{\relax}
\providecommand{\bibinfo}[2]{#2}
\providecommand\BIBentrySTDinterwordspacing{\spaceskip=0pt\relax}
\providecommand\BIBentryALTinterwordstretchfactor{4}
\providecommand\BIBentryALTinterwordspacing{\spaceskip=\fontdimen2\font plus
\BIBentryALTinterwordstretchfactor\fontdimen3\font minus \fontdimen4\font\relax}
\providecommand\BIBforeignlanguage[2]{{%
\expandafter\ifx\csname l@#1\endcsname\relax
\typeout{** WARNING: IEEEtran.bst: No hyphenation pattern has been}%
\typeout{** loaded for the language `#1'. Using the pattern for}%
\typeout{** the default language instead.}%
\else
\language=\csname l@#1\endcsname
\fi
#2}}

\bibitem{Biswal2017}
P.~Biswal and P.~K. Mohanty, ``Development of quadruped walking robots: A review,'' \emph{Ain Shams Engineering Journal}, vol.~12, no.~2, pp. 2017--2031, 2021.

\bibitem{lunardi2024}
G.~{Lunardi}, T.~{Corb{\`e}res}, C.~{Mastalli}, N.~{Mansard}, T.~{Flayols}, S.~{Tonneau}, and A.~D. {Prete}, ``{Reference-Free Model Predictive Control for Quadrupedal Locomotion},'' \emph{IEEE Access}, vol.~12, pp. 689--698, 2024.

\bibitem{dicarlo2018}
J.~Di~Carlo, P.~M. Wensing, B.~Katz, G.~Bledt, and S.~Kim, ``Dynamic locomotion in the mit cheetah 3 through convex model-predictive control,'' in \emph{2018 IEEE/RSJ International Conference on Intelligent Robots and Systems (IROS)}, 2018, pp. 1--9.

\bibitem{Rudin2022}
N.~Rudin, D.~Hoeller, M.~Bjelonic, and M.~Hutter, ``Advanced skills by learning locomotion and local navigation end-to-end,'' in \emph{2022 IEEE/RSJ International Conference on Intelligent Robots and Systems (IROS)}, 2022, pp. 2497--2503.

\bibitem{omar2023}
S.~Omar, L.~Amatucci, V.~Barasuol, G.~Turrisi, and C.~Semini, ``Safesteps: Learning safer footstep planning policies for legged robots via model-based priors,'' in \emph{2023 IEEE-RAS 22nd International Conference on Humanoid Robots (Humanoids)}, 2023, pp. 1--8.

\bibitem{howell2022}
T.~Howell, N.~Gileadi, S.~Tunyasuvunakool, K.~Zakka, T.~Erez, and Y.~Tassa, ``{Predictive Sampling: Real-time Behaviour Synthesis with MuJoCo},'' 2022.

\bibitem{Stulp2012}
F.~Stulp and O.~Sigaud, ``Path integral policy improvement with covariance matrix adaptation,'' \emph{ArXiv}, vol. abs/1206.4621, 2012.

\bibitem{Bradbury2018}
J.~Bradbury, R.~Frostig, P.~Hawkins, M.~J. Johnson, C.~Leary, D.~Maclaurin, G.~Necula, A.~Paszke, J.~Vander{P}las, S.~Wanderman-{M}ilne, and Q.~Zhang, ``{JAX}: composable transformations of {P}ython+{N}um{P}y programs,'' 2018.

\bibitem{amatucci}
L.~Amatucci, J.-H. Kim, J.~Hwangbo, and H.-W. Park, ``Monte carlo tree search gait planner for non-gaited legged system control,'' in \emph{2022 International Conference on Robotics and Automation (ICRA)}, 2022, pp. 4701--4707.

\bibitem{kang2022}
D.~Kang, F.~De~Vincenti, N.~C. Adami, and S.~Coros, ``Animal motions on legged robots using nonlinear model predictive control,'' in \emph{2022 IEEE/RSJ International Conference on Intelligent Robots and Systems (IROS)}, 2022, pp. 11\,955--11\,962.

\bibitem{Ruben2023}
R.~Grandia, F.~Jenelten, S.~Yang, F.~Farshidian, and M.~Hutter, ``Perceptive locomotion through nonlinear model-predictive control,'' \emph{IEEE Transactions on Robotics}, vol.~39, no.~5, pp. 3402--3421, 2023.

\bibitem{qpOASES}
H.~J. Ferreau, C.~Kirches, A.~Potschka, H.~G. Bock, and M.~Diehl, ``qpoases: a parametric active-set algorithm for quadratic programming,'' \emph{Mathematical Programming Computation}, vol.~6, pp. 327 -- 363, 2014.

\bibitem{HPIPM}
G.~Frison and M.~Diehl, ``Hpipm: a high-performance quadratic programming framework for model predictive control,'' \emph{IFAC-PapersOnLine}, vol.~53, no.~2, pp. 6563--6569, 2020, 21st IFAC World Congress.

\bibitem{cheng2023}
X.~Cheng, K.~Shi, A.~Agarwal, and D.~Pathak, ``Extreme parkour with legged robots,'' \emph{arXiv:2309.14341}, 2023.

\bibitem{zhuang2023}
Z.~Zhuang, Z.~Fu, J.~Wang, C.~Atkeson, S.~Schwertfeger, C.~Finn, and H.~Zhao, ``Robot parkour learning,'' in \emph{Conference on Robot Learning ({CoRL})}, 2023.

\bibitem{asmar2022}
D.~M. Asmar, R.~Senanayake, S.~Manuel, and M.~J. Kochenderfer, ``Model predictive optimized path integral strategies,'' \emph{2023 IEEE International Conference on Robotics and Automation (ICRA)}, pp. 3182--3188, 2022.

\bibitem{Pravitra2021}
J.~Pravitra, E.~Theodorou, and E.~N. Johnson, ``Flying complex maneuvers with model predictive path integral control,'' in \emph{AIAA Scitech 2021 Forum}.\hskip 1em plus 0.5em minus 0.4em\relax American Institute of Aeronautics and Astronautics, 2021.

\bibitem{Williams2017}
G.~Williams, P.~Drews, B.~Goldfain, J.~M. Rehg, and E.~A. Theodorou, ``Information-theoretic model predictive control: Theory and applications to autonomous driving,'' \emph{IEEE Transactions on Robotics}, vol.~34, pp. 1603--1622, 2017.

\bibitem{Yang2019}
Y.~Yang, K.~Caluwaerts, A.~Iscen, T.~Zhang, J.~Tan, and V.~Sindhwani, ``Data efficient reinforcement learning for legged robots,'' in \emph{Conference on Robot Learning ({CoRL})}, 2019.

\bibitem{hutter_sampling}
J.~Carius, R.~Ranftl, F.~Farshidian, and M.~Hutter, ``Constrained stochastic optimal control with learned importance sampling: A path integral approach,'' \emph{The International Journal of Robotics Research}, vol.~41, pp. 189 -- 209, 2021.

\bibitem{Modugno2017}
V.~Modugno, G.~Nava, D.~Pucci, F.~Nori, G.~Oriolo, and S.~Ivaldi, ``Safe trajectory optimization for whole-body motion of humanoids,'' in \emph{2017 IEEE-RAS 17th International Conference on Humanoid Robotics (Humanoids)}, 2017, pp. 763--770.

\bibitem{Theodorou2010}
E.~Theodorou, J.~Buchli, and S.~Schaal, ``A generalized path integral control approach to reinforcement learning.'' \emph{Journal of Machine Learning Research}, vol.~11, pp. 3137--3181, 2010.

\bibitem{winkler}
A.~W. Winkler, C.~D. Bellicoso, M.~Hutter, and J.~Buchli, ``Gait and trajectory optimization for legged systems through phase-based end-effector parameterization,'' \emph{IEEE Robotics and Automation Letters}, vol.~3, no.~3, pp. 1560--1567, 2018.

\bibitem{Rathod2021}
N.~Rathod, A.~Bratta, M.~Focchi, M.~Zanon, O.~Villarreal, C.~Semini, and A.~Bemporad, ``Model predictive control with environment adaptation for legged locomotion,'' \emph{IEEE Access}, vol.~9, pp. 145\,710--145\,727, 2021.

\bibitem{NMPC_contact}
M.~{Neunert}, M.~{Stäuble}, M.~{Giftthaler}, C.~D. {Bellicoso}, J.~{Carius}, C.~{Gehring}, M.~{Hutter}, and J.~{Buchli}, ``Whole-body nonlinear model predictive control through contacts for quadrupeds,'' \emph{IEEE Robotics and Automation Letters}, vol.~3, no.~3, pp. 1458--1465, 2018.

\bibitem{different_gait}
C.~D. Bellicoso, F.~Jenelten, C.~Gehring, and M.~Hutter, ``Dynamic locomotion through online nonlinear motion optimization for quadrupedal robots,'' \emph{IEEE Robotics and Automation Letters}, vol.~3, no.~3, pp. 2261--2268, 2018.

\bibitem{Boussema}
C.~Boussema, M.~J. Powell, G.~Bledt, A.~J. Ijspeert, P.~M. Wensing, and S.~Kim, ``Online gait transitions and disturbance recovery for legged robots via the feasible impulse set,'' \emph{IEEE Robotics and Automation Letters}, vol.~4, no.~2, pp. 1611--1618, 2019.

\bibitem{Verschueren2019}
R.~Verschueren, G.~Frison, D.~Kouzoupis, J.~Frey, N.~van Duijkeren, A.~Zanelli, B.~Novoselnik, T.~Albin, R.~Quirynen, and M.~Diehl, ``acados—a modular open-source framework for fast embedded optimal control,'' \emph{Mathematical Programming Computation}, vol.~14, pp. 147 -- 183, 2019.

\bibitem{todorov2012mujoco}
E.~Todorov, T.~Erez, and Y.~Tassa, ``Mujoco: A physics engine for model-based control,'' in \emph{2012 IEEE/RSJ International Conference on Intelligent Robots and Systems (IROS)}, 2012, pp. 5026--5033.

\end{thebibliography}

\end{document}